\documentclass{INTERSPEECH2023}

\interspeechcameraready

\usepackage{graphicx}
\usepackage[caption=false]{subfig}
\usepackage{multicol}
\usepackage{xcolor}

\usepackage{multirow}
\usepackage{makecell}

\usepackage{amsmath}
\usepackage{amsfonts}

\title{Relation-based Counterfactual Data Augmentation and Contrastive Learning for Robustifying Natural Language Inference Models}
\name{Heerin Yang$^{1,3}$\thanks{$^*$This work was done while Heerin Yang was at Sogang University. $^\dagger$corresponding author.}, Seung-won Hwang$^2$, Jungmin So$^{1\dagger}$}

\address{
  $^1$Dept. of Computer Science and Engineering, Sogang University, Korea \\
  $^2$Dept. of Computer Science and Engineering, Seoul National University, Korea \\
  $^3$LG Electronics, Korea
  }
\email{heerin.yang@lge.com, seungwonh@snu.ac.kr, jso1@sogang.ac.kr}

\begin{document}

\maketitle
 
\begin{abstract}
Although pre-trained language models show good performance on various natural language processing tasks, they often rely on non-causal features and patterns to determine the outcome. For natural language inference tasks, previous results have shown that even a model trained on a large number of data fails to perform well on counterfactually revised data, indicating that the model is not robustly learning the semantics of the classes. In this paper, we propose a method in which we use token-based and sentence-based augmentation methods to generate counterfactual sentence pairs that belong to each class, and apply contrastive learning to help the model learn the difference between sentence pairs of different classes with similar contexts. Evaluation results with counterfactually-revised dataset and general NLI datasets show that the proposed method can improve the performance and robustness of the NLI model.
\end{abstract}

\section{Introduction}

A recently popular approach to solving natural language processing (NLP) problems is to use a pre-trained language model such as BERT \cite{devlin-etal-2019-bert} and RoBERTa \cite{liu2019roberta}, then fine-tune the model on a downstream task such as text classification. Although trained models achieve outstanding performance in various tasks such as sentiment analysis \cite{maas-etal-2011-learning, socher-etal-2013-recursive} and natural language inference (NLI) \cite{bowman-etal-2015-large, williams-etal-2018-broad}, it is well-known that these models often make decisions based on spurious patterns and correlations and therefore do not generalize well to other datasets. For example, NLI classifiers may learn that a sentence pair having significant lexical overlap is a sign that they are in an entailment relationship, which is not necessarily true \cite{mccoy-etal-2019-right}.

Kaushik et al. \cite{kaushik2020learning} showed that a model trained on the original dataset performs poorly on a counterfactually revised dataset, which is another evidence that the model is relying on spurious patterns to classify data. For collecting counterfactually revised data, human workers were asked to edit given data samples to produce new samples that have different labels than the original ones.
For example, if the given NLI sentence pair is ``A man in a boom lift bucket welds. A man is working. (entailment)'', then the worker writes counterfactual samples by revising the premise such as ``A woman in a boom lift bucket welds. A man is working (contradiction)'' or ``A person in a boom lift bucket welds. A man is working. (neutral)''. A classifier trained on the original dataset classifies all three pairs as entailment.

In this paper, we consider automatically generating counterfactual data for NLI tasks. While Kaushik et al. \cite{kaushik2020learning} claims that the counterfactually-revised train sets by human workers could improve model performance on the challenge sets, human annotation is costly. Our goal is to make the NLI model more robust to counterfactually revised data without getting help from human annotators.
Compared to other NLP tasks where a single sentence or passage is considered as input, NLI poses a unique challenge where a sentence pair is given as input and its relation is an important feature for classification. However, existing augmentation methods such as EDA \cite{wei-zou-2019-eda} regards an NLI sentence pair as a single unit of input without considering their relation. In contrast, 
we counterfactually augment hypothesis sentences for a fixed premise and vice versa, and represent their relation more explicitly, as a distance, to minimize or maximize during contrastive learning.

Specifically, we apply contrastive learning with the generated set, pulling the original pair and the generated pair with the same label together while pushing the original pair and other generated pairs away, in the embedding space. We empirically find that this method is more effective than applying supervised contrastive learning with unrelated sentence pairs \cite{NEURIPS2020_d89a66c7}. There are other recent methods \cite{ng-etal-2020-ssmba, Moon_Mo_Lee_Lee_Shin_2021, c2l2022} using automatic data augmentation and contrastive learning to make the model more robust, but their improvements are limited mostly because they do not consider the unique characteristics of NLI where the inputs are pairs and their relations are important.
The experimental results show that the proposed method achieves better accuracy compared to other robust text classification methods on counterfactually revised NLI datasets \cite{kaushik2020learning} as well as general NLI datasets\footnote{The codes are available at https://github.com/hryang06/rda-rcl.}.

\begin{figure*}[tb]
\centering
\subfloat[Process of token-level data augmentation. The word ``musician'' is replaced to ``guitarist'', and the label is changed to ``neutral''.\label{fig:token-level}]{\includegraphics[width=0.9\textwidth]{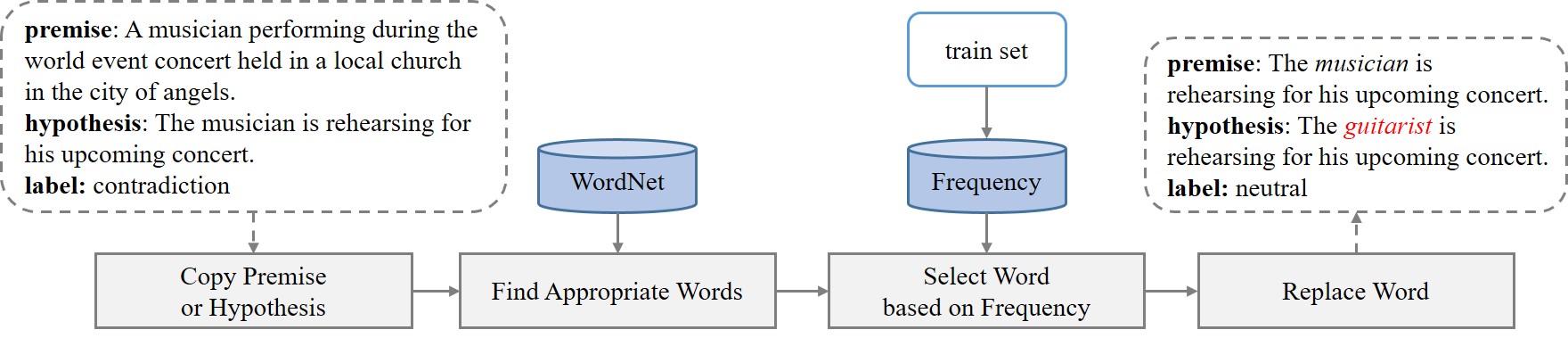}}
\hfill
\subfloat[Process of sentence-level data augmentation. The classifier and the generator are iteratively trained with augmented data. \label{fig:sentence-level}]{\includegraphics[width=0.85\textwidth]{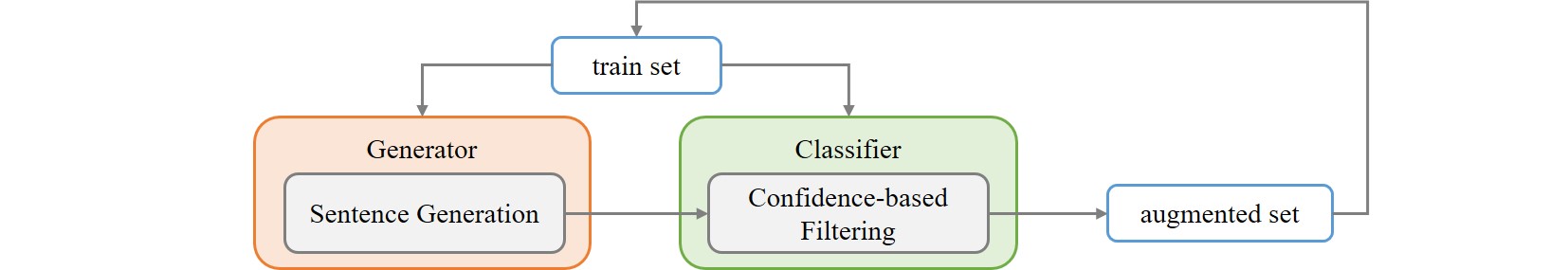}}
\caption{Our proposed data augmentation framework.}
\label{fig:da-generation}
\end{figure*}

\section{Related work}

Data augmentation for NLP tasks can be divided into token-level and sentence-level augmentation. Token-level augmentation modifies individual words, such as substituting a word with synonyms \cite{wang-yang-2015-thats, 10.5555/2969239.2969312}, randomly inserting, deleting, or swapping tokens \cite{wei-zou-2019-eda}. Language models can be used for augmentation, by masking a particular word and using the model to fill in the blank \cite{kobayashi-2018-contextual, gao-etal-2019-soft}. The quality of token-based augmentation depends on selecting which token to insert or remove, such as finding the rationale tokens and replacing them \cite{ng-etal-2020-ssmba, Moon_Mo_Lee_Lee_Shin_2021, c2l2022}.
Sentence-level augmentation generates an entire sentence rather than modifying tokens from the original text. Examples include back-translation \cite{edunov-etal-2018-understanding}, paraphrasing \cite{iyyer-etal-2018-adversarial}, and conditional generation \cite{Anaby-Tavor_Carmeli_Goldbraich_Kantor_Kour_Shlomov_Tepper_Zwerdling_2020}.
While sentence-level augmentation can generate more diverse text compared to token-level augmentation, it is more difficult to assign labels or determine the quality of generated data. Therefore, filtering methods based on teacher models are often used to select good quality data \cite{wu-etal-2022-generating}.

Contrastive learning is recently recognized as an effective method to improve model performance \cite{he2019moco, chen2020simple}. It is shown to make the model more robust to perturbations and improve its generalization ability.
In unsupervised contrastive learning, an original input is paired with a slightly modified input to form a positive pair and paired with a different sample to form a negative pair. 
For example, in C\textsuperscript{2}L \cite{c2l2022}, a negative pair is created by masking keyword tokens from the original text, while a positive pair is created by masking non-keyword tokens. 
It is also possible to use contrastive learning in a supervised learning context, gathering same-class samples together in the feature space, while separating different-class samples \cite{NEURIPS2020_d89a66c7}.

\section{Proposed Method}

\subsection{Relation-based Counterfactual Data Augmentation}
In our proposed method, we first generate a set of entailment, neutral, and contradiction sentence pairs for each sentence pair in the train set. We apply two major data augmentation approaches, token-level and sentence-level augmentation, tailored for NLI tasks to generate factual and counterfactual data. 

\subsubsection{Token-level Data Augmentation}
While simple methods such as synonym replacement \cite{wei-zou-2019-eda} can be used to generate class-preserving data, it is not trivial to generate counterfactual data.
Suppose the original premise-hypothesis pair is ``A man is walking down the street. A man is outside walking. (entailment)'' Changing the hypothesis to ``A woman is outside walking.'' will make the relation contradictory. However, if the original premise was ``A person is walking down the street.'', changing the hypothesis as such will not alter the label (neutral).
In our proposed method, we take only one sentence from the original pair and copy the sentence to make an entailment pair (e.g. ``A man is outside walking. A man is outside walking.''). From this pair, we apply word substitution on either premise or hypothesis to generate sentence pairs that belong to the three classes. 

Figure \ref{fig:token-level} shows our token-level data augmentation process.
We first choose a random noun word in the sentence using \texttt{spaCy}\footnote{https://spacy.io}. Then, we use \texttt{WordNet}\footnote{https://wordnet.princeton.edu} to find the substitution words. Table \ref{tab:da-method} shows how the substitution words are selected based on the revised sentence and the target class. For example, we choose a synonym or a hypernym to make an entailment sentence, a hyponym to make a neutral sentence, and an antonym or co-hyponym to make a contradiction sentence. Among candidate words, we sample a word based on its frequency in the train set. In the case where no candidate substitution is found, the sentence pair is omitted from contrastive learning.
Table \ref{tab:da-examples} shows the sentences generated by four different configurations.
One limitation of our scheme is that we only substitute nouns in the sentence. Substituting words other than nouns for counterfactual data generation is left for future work.


\begin{table}[ht]
\centering
\begin{tabular}{ccc}
\Xhline{3\arrayrulewidth}
Target Label    & Revise Premise    & Revise Hypothesis \\ 
\hline
entailment      & synonym, hyponym  & synonym, hypernym \\
neutral         & hypernym          & hyponym \\
contradiction   & \multicolumn{2}{c}{antonym, co-hyponym} \\ 
\Xhline{3\arrayrulewidth}
\end{tabular}
\caption{Relation types used in word substitution to generate a sample of the target label.}
\label{tab:da-method}
\end{table}

\vspace{-7mm}


\subsubsection{Sentence-level Data Augmentation}
Conditional generation techniques such as LAMBADA \cite{Anaby-Tavor_Carmeli_Goldbraich_Kantor_Kour_Shlomov_Tepper_Zwerdling_2020} can be used to generate the hypothesis sentence conditioned on the premise sentence (and the label), and vice versa. We follow the basic approach of LAMBADA, but instead of generating independent samples, we let the generator create a set of entailment, neutral, and contradiction sentences for each input sentence.

Pre-trained sequence-to-sequence language models such as GPT-2 \cite{radford2019language}, BART \cite{lewis-etal-2020-bart}, and T5 \cite{2020t5} can be used as a sentence generator, and we use T5 model to generate counterfactual premise or hypothesis sentences. The problem with using a generator model is that the generated sentence pairs may have incorrect labels. A typical method to address this problem is to evaluate the generated data samples on a classifier model trained on the original data, and filter out samples that have low confidence in the target class \cite{ng-etal-2020-ssmba, wu-etal-2022-generating}. 

\begin{table*}[ht]
\centering
\includegraphics[width=\textwidth]{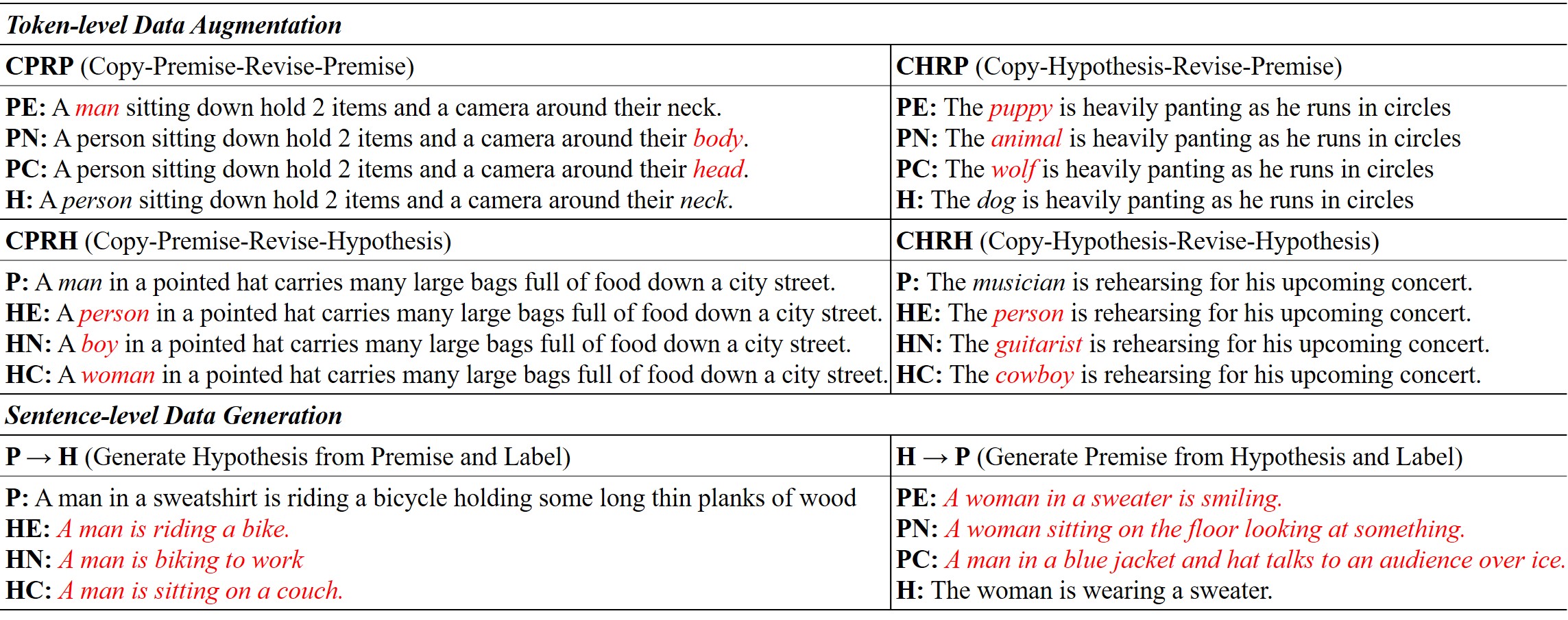}
\caption{Counterfactual data generated using our data augmentation methods. Replaced or generated words are marked in red.}
\label{tab:da-examples}
\end{table*}

\begin{figure}[ht]
\centering
\includegraphics[width=0.38\textwidth]{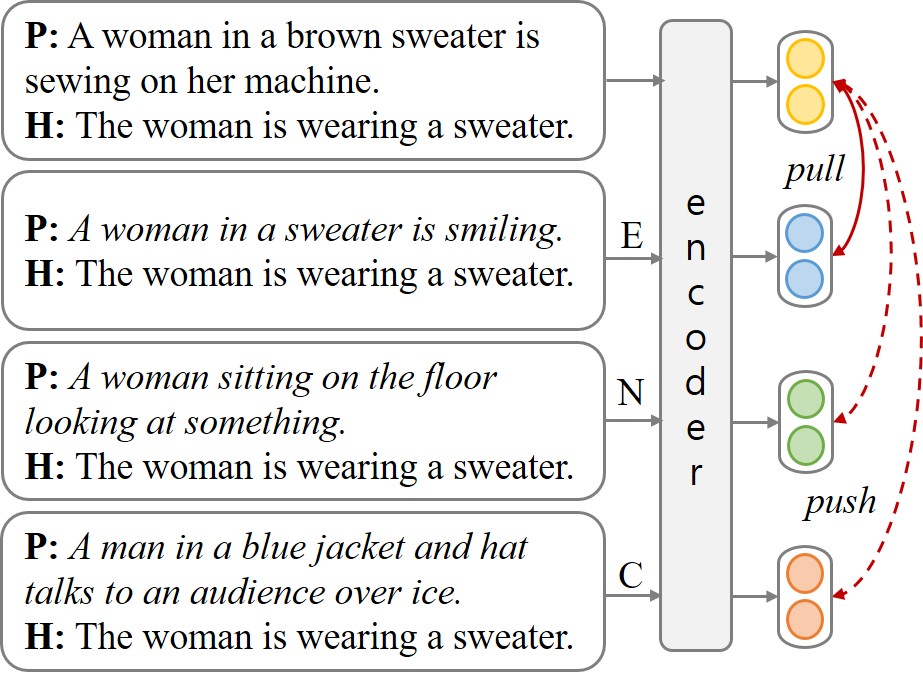}
\caption{Contrastive learning with augmented data.}
\label{fig:cl-frame}
\end{figure}


Figure \ref{fig:sentence-level} illustrates the sentence-level augmentation process. We first train a classifier model and a generator model with the original train set. Then, the generator model generates three sentences for each sentence pair in the original set. For the generated sentence pairs, we apply confidence-based filtering and drop samples with model confidence lower than a threshold $\tau$. We go through an iterative process where the augmented set becomes the train set which is used to train the classifier and the generator. This iterative process goes on until we have obtained a full set (entailment, neutral, contradiction, plus original pair) for over 95\% of samples in the original set. The samples that could not construct a full set during the augmentation stage are omitted from relation-based contrastive learning. Table \ref{tab:da-examples} shows the sentences generated using our method.


\subsection{Relation-based Contrastive Learning}
Once each sentence pair is augmented with sentence pairs corresponding to all three classes, we train the classifier with the augmented set. The model is first trained with the contrastive learning objective. A set of four sentence pairs (original, entailment, neutral, contradiction) is passed through the encoder to obtain the sentence embedding vectors. Then, cosine similarity is measured between the original embedding vector and the embedding vectors of other sentence pairs. Finally, the contrastive loss $\mathcal{L}_{CL}$ is calculated according to Eq. \ref{eq:1}.

\vspace{-1mm}

\begin{equation}
\mathcal{L}_{CL} = -log \cfrac{exp(sim(x, x_y) / T)}{\sum_{c=0}^C exp(sim(x, x_c) / T)}
\label{eq:1}
\end{equation}

The contrastive learning process is shown in Figure \ref{fig:cl-frame}. Suppose the original label is entailment. Then, the distance between the embedding vectors of the original and entailment pair is minimized, while the distances between the embedding vectors of the original and other pairs are maximized. After contrastive learning, the model is trained using cross-entropy loss. 

\section{Experimental Results}

\begin{table*}[ht]
\centering
\begin{tabular}{l>{\centering}m{0.1\textwidth}>{\centering}m{0.1\textwidth}>{\centering}m{0.1\textwidth}>{\centering\arraybackslash}m{0.1\textwidth}}
\Xhline{3\arrayrulewidth}
\multicolumn{1}{c}{\multirow{2}{*}{Model}} & \multicolumn{4}{c}{CF-SNLI} \\ \cline{2-5} 
\multicolumn{1}{c}{}    & Original  & RP & RH & RP \& RH \\ 
\hline
BERT-base               & 75.5 \small$\pm 1.4$ & 41.8 \small$\pm 2.6$ & 64.5 \small$\pm 2.0$ & 53.1 \small$\pm 2.2$ \\
+ SSMBA \cite{ng-etal-2020-ssmba} \textsuperscript{\textbf{*}}      & 75.8 \small$\pm 1.5$ & 42.5 \small$\pm 0.9$ & 65.0 \small$\pm 0.3$ & 53.8 \small$\pm 0.5$ \\ 
+ MCL (grad+SL) \cite{Moon_Mo_Lee_Lee_Shin_2021} \textsuperscript{\textbf{*}} & 78.3 \small$\pm 1.1$ & 40.0 \small$\pm 1.3$ & 64.5 \small$\pm 1.3$ & 52.2 \small$\pm 1.3$ \\ 
+ C\textsuperscript{2}L \cite{c2l2022} \textsuperscript{\textbf{*}} & 76.2 \small$\pm 1.7$ & 43.1 \small$\pm 2.5$ & 65.8 \small$\pm 1.7$ & 54.5 \small$\pm 2.1$ \\ 
+ SCL                       & 75.7 \small$\pm 1.1$    & 42.3 \small$\pm 1.2$ & 65.9 \small$\pm 0.9$ & 54.1 \small$\pm 1.0$ \\
+ RDA \textit{(Ours)}       & 77.7 \small$\pm 1.1$ & 46.5 \small$\pm 0.8$ & 67.3 \small$\pm 1.6$ & 56.9 \small$\pm 1.0$ \\
+ RDA-RCL \textit{(Ours)}   & \textbf{79.3 \small$\pm 1.0$} & \textbf{47.5 \small$\pm 0.9$} & \textbf{68.0 \small$\pm 0.5$} & \textbf{57.8 \small$\pm 0.6$} \\
\hline
RoBERTa-base                & 81.4 \small$\pm 1.9$ & 51.5 \small$\pm 0.5$ & 68.2 \small$\pm 1.4$ & 59.8 \small$\pm 0.9$ \\
+ SCL                       & 82.0 \small$\pm 1.2$ & 51.5 \small$\pm 0.7$ & 68.6 \small$\pm 1.3$ & 60.1 \small$\pm 1.0$ \\
+ RDA \textit{(Ours)}       & 84.5 \small$\pm 1.0$ & 59.3 \small$\pm 0.8$ & 73.3 \small$\pm 0.8$ & 66.3 \small$\pm 0.4$ \\
+ RDA-RCL \textit{(Ours)}   & \textbf{84.7 \small$\pm 1.3$} & \textbf{59.6 \small$\pm 0.8$} & \textbf{73.6 \small$\pm 0.4$} & \textbf{66.6 \small$\pm 0.6$} \\
\Xhline{3\arrayrulewidth}
\end{tabular}
\caption{\label{tab:main-acc} Accuracy of methods on counterfactually-augmented SNLI dataset. \textbf{\textsuperscript{*}}results from Choi et al. \cite{c2l2022}.}
\end{table*}

\subsection{Experiment Setup}

We use the counterfactually augmented SNLI dataset (CF-SNLI), which is also used by previous works for testing the robustness of NLI models \cite{ng-etal-2020-ssmba, Moon_Mo_Lee_Lee_Shin_2021, c2l2022}. CF-SNLI set contains ``original'' train and test sets sampled from SNLI \cite{bowman-etal-2015-large}. It also has ``revised premise'' (RP) and ``revised hypothesis'' (RH) set, where the premise and hypothesis sentences are revised by human workers to produce sentence pairs with relations other than the original pair. 
We evaluate with all CF-SNLI test sets and also general NLI datasets-- SNLI test set, MNLI dev-matched set, and MNLI dev-mismatched set \cite{williams-etal-2018-broad}.

We use BERT (bert-base-uncased) and RoBERTa (roberta-base) as pre-trained language models. For BERT, the model is trained with contrastive loss for 10 epochs (lr=1e-5), followed by cross-entropy loss for 3 epochs (lr=3e-5). For RoBERTa, the model is trained with contrastive loss for 10 epochs (lr=2e-6), followed by cross-entropy loss for 5 epochs (lr=1e-5). We use 0.1 as the temperature $T$ in Eq. \ref{eq:1}. In sentence generation, the threshold $\tau$ is empirically tuned to 0.9. The results were not sensitive to $\tau$, unless we choose a very low number.

We compare the performance of our method with other recent methods based on counterfactual data. SSMBA \cite{ng-etal-2020-ssmba} uses a corruption function to perturb the original text and a reconstruction function to generate a new text in the underlying data manifold. MASKER \cite{Moon_Mo_Lee_Lee_Shin_2021} selects keywords in the text using attention scores or gradients and applies masked keyword reconstruction to help the model learn the context rather than relying on particular tokens. C\textsuperscript{2}L \cite{c2l2022} generates factual and counterfactual samples by masking non-causal and causal tokens in the original text, and applies contrastive learning to help the model learn to rely on causal tokens.

\begin{table}[t]
\centering
\begin{tabular}{lccc}
\Xhline{3\arrayrulewidth}
\multicolumn{1}{c}{\multirow{2}{*}{Model}} & \multicolumn{1}{c}{SNLI} & \multicolumn{2}{c}{MNLI} \\ \cline{2-2} \cline{3-4}
\multicolumn{1}{c}{}    & test  & dev-m & dev-mm \\ 
\hline
BERT-base       & \small76.0 $\pm 0.7$ & \small52.1 $\pm 2.5$ & \small51.8 $\pm 3.1$ \\
+ SCL           & \small75.8 $\pm 0.5$ & \small52.8 $\pm 2.0$ & \small53.4 $\pm 2.5$ \\
+ RDA           & \small76.7 $\pm 0.4$ & \small59.5 $\pm 0.9$ & \small60.5 $\pm 1.6$ \\ 
+ RDA-RCL       & \small\textbf{77.8 $\pm 0.6$} & \small\textbf{60.1 $\pm 2.1$} & \small\textbf{61.5 $\pm 2.9$} \\
\hline
RoBERTa-base    & \small79.7 $\pm 0.9$ & \small58.5 $\pm 3.3$ & \small60.0 $\pm 3.8$ \\
+ SCL           & \small80.2 $\pm 1.2$ & \small59.9 $\pm 2.8$ & \small62.0 $\pm 3.1$ \\
+ RDA           & \small83.1 $\pm 0.2$ & \small69.7 $\pm 0.1$ & \small70.8 $\pm 0.4$ \\
+ RDA-RCL       & \small\textbf{83.1 $\pm 0.4$} & \small\textbf{70.5 $\pm 0.3$} & \small\textbf{71.6 $\pm 0.5$} \\
\Xhline{3\arrayrulewidth}
\end{tabular}
\caption{\label{tab:nli-acc} Accuracy on SNLI test and MNLI dev sets.}
\end{table}

\subsection{Results}
In the tables, BERT-base and RoBERTa-base are baseline models fine-tuned with CF-SNLI original train set, and SCL refers to supervised contrastive learning \cite{NEURIPS2020_d89a66c7}, where contrastive learning is applied without data augmentation. RDA (Relation-based Data Augmentation) and RCL (Relation-based Contrastive Learning) are the components of our proposed method. RDA is the case where only data augmentation is applied, whereas RDA-RCL is the case where contrastive learning is also applied. We seek to answer the following research questions.

\noindent{\textit{\textbf{RQ1: Does the proposed method perform better than the baseline and other data augmentation methods?}}} In Table \ref{tab:main-acc}, models trained with different methods were evaluated on CF-SNLI test sets. We can observe that the proposed method achieves higher accuracy over the baseline and other methods in all sets for both BERT and RoBERTa models. The performance improvement is 6-8\% for the RP set and 3-5\% for the RH set, respectively. The proposed method also achieves 3-4\% improvement over the baseline on the original test set, which indicates that the method not only improves robustness to counterfactual revisions but helps the model performance in general.

\noindent{\textit{\textbf{RQ2: Does the proposed method show good performance on the general NLI sets?}}} Since it is important to see whether the proposed method is effective in datasets other than CF-SNLI, we have evaluated the models on SNLI test set and MNLI dev sets. Since CF-SNLI original set is sampled from SNLI, we can say that SNLI is an in-domain set whereas MNLI is an out-of-domain set. Table \ref{tab:nli-acc} shows that the proposed method achieves significantly higher accuracy over baseline for both BERT and RoBERTa models. While the accuracy improvement is 2-4\% for SNLI, our method achieves 8-12\% higher accuracy over baseline on MNLI dev sets, which shows that the method is also effective in improving generalization performance.



\noindent{\textit{\textbf{RQ3: Is the proposed method better than general supervised contrastive learning?}}} The relation-based contrastive learning applies supervised contrastive learning on sentence pairs with the common premise or hypothesis. The question is whether it is better than applying general SCL where contrastive learning is applied to different sentence pairs. Table \ref{tab:main-acc} and \ref{tab:nli-acc} show that applying general SCL only achieves marginal improvement over baseline, while RDA-RCL shows significantly better results for different datasets as well as different models.

\noindent{\textit{\textbf{RQ4: Does applying relation-based contrastive learning helps improving model performance?}}} Since we assign labels to counterfactually generated sentence pairs, augmenting them to the train set already helps improve model performance. However, applying relation-based contrastive learning further boosts performance. In Table \ref{tab:main-acc} and \ref{tab:nli-acc}, RDA-RCL achieves up to 2\% higher accuracy over RDA for varying datasets and models, while there is no case where RCL degrades the performance.

Overall, the proposed method is an effective way to robustify NLI models against counterfactual revisions, as well as improve model accuracy and generalization performance.

\section{Conclusions}

This paper studied the effectiveness of relation-based data augmentation and contrastive learning on NLI tasks. 
For a given sentence pair, the proposed method applies token-based and sentence-based augmentation to generate a set of counterfactual sentence pairs for all classes. 
Relation-based contrastive learning is done using the set of counterfactual sentence pairs to help the model effectively learn the difference between classes. 
Empirical results show that our methods can improve the robustness of classifier models on NLI tasks. 
Since any sentences can be used as input to our methods, a possible future work can use our methods to create a large number of NLI sentence pairs using inputs outside the train set.

\section{Acknowledgements}

This work was supported by the NRF (National Research Foundation) of Korea under grant no. 2021S1A5A2A03064795.

\newpage

\bibliographystyle{IEEEtran}
\bibliography{mybib}

\end{document}